\documentclass[lettersize,journal]{IEEEtran}
\usepackage{amsmath,amsfonts}
\usepackage{algorithmic}
\usepackage{algorithm}
\usepackage{array}
\usepackage{textcomp}
\usepackage{stfloats}
\usepackage{url}
\usepackage{verbatim}
\usepackage{graphicx}
\usepackage{cite}
\usepackage[colorlinks,linkcolor=black]{hyperref}
\usepackage{caption}
\usepackage{soul}
\usepackage{xcolor}
\usepackage{multirow,epstopdf,subfigure}
\usepackage{booktabs} % 解决 \toprule, \midrule, \bottomrule 报错
\usepackage{xcolor}   % 若使用 \textcolor
\usepackage{tabularx} % 若使用 tabularx 自适应表格宽度

\begin{document}

\title{Inter-Image Pixel Shuffling for Multi-focus Image Fusion}

\author{IEEE Publication Technology,~\IEEEmembership{Staff,~IEEE,}
        % <-this % stops a space
\thanks{This paper was produced by the IEEE Publication Technology Group. They are in Piscataway, NJ.}% <-this % stops a space
\thanks{Manuscript received April 19, 2021; revised August 16, 2021.}}

\author{
    Huangxing Lin, Rongrong Ma, Cheng Wang
    
%    \thanks{This work was supported in part by the Fund of the National Key Laboratory of Automatic Target Recognition under Grant 230101; in part by the Project funded by China Post-Doctoral Science Foundation under Grant Grant 2023M744322; in part by the Post-Doctoral Fellowship Program of China Post-Doctoral Science Foundation under Grant GZC20233543  (e-mail: linhuangxing23@nudt.edu.cn). }
    
    \thanks{Huangxing Lin, Rongrong Ma and Cheng Wang are with the College of Computer Science and Technology, Huaqiao University, Xiamen 361021, China (e-mail: linhuangxing@hqu.edu.cn).}
	}
 
%\thanks{This work was supported in part by the Fund of the National Key Laboratory of Automatic Target Recognition under Grant 230101; in part by the Project funded by China Post-Doctoral Science Foundation under Grant Grant 2023M744322; in part by the Post-Doctoral Fellowship Program of China Post-Doctoral Science Foundation under Grant GZC20233543. }
%
%\thanks{Yongxiang Liu, Ruikang Hu, Qinmu Shen and Huangxing Lin are with the College of Electronic Science, National University of Defense Technology, Changsha, 410073, China (e-mail: linhuangxing23@nudt.edu.cn).}

% The paper headers
\markboth{}%
{Shell \MakeLowercase{\textit{et al.}}: A Sample Article Using IEEEtran.cls for IEEE Journals}

\maketitle

\begin{abstract}
	Multi-focus image fusion aims to combine multiple partially focused images into a single all-in-focus image. Although deep learning has shown promise in this task, its effectiveness is often limited by the scarcity of suitable training data. This paper introduces Inter-image Pixel Shuffling (IPS), a novel method that allows neural networks to learn multi-focus image fusion without requiring actual multi-focus images. IPS reformulates the task as a pixel-wise classification problem, where the goal is to identify the focused pixel from a pixel group at each spatial position. In this method, pixels from a clear optical image are treated as focused, while pixels from a low-pass filtered version of the same image are considered defocused. By randomly shuffling the focused and defocused pixels at identical spatial positions in the original and filtered images, IPS generates training data that preserves spatial structure while mixing focus-defocus information. The model is trained to select the focused pixel from each spatially aligned pixel group, thus learning to reconstruct an all-in-focus image by aggregating sharp content from the input. To further enhance fusion quality, IPS adopts a cross-image fusion network that integrates the localized representation power of convolutional neural networks with the long-range modeling capabilities of state space models. This design effectively leverages both spatial detail and contextual information to produce high-quality fused results. Experimental results indicate that IPS significantly outperforms existing multi-focus image fusion methods, even without training on multi-focus images.

\end{abstract}


\section{Introduction}
\IEEEPARstart 
DUE to the depth-of-field limitations of optical lenses, only objects within a certain distance from the camera can be sharply imaged, while regions outside the focal plane appear blurred. To address this issue, multi-focus image fusion (MFIF) algorithms are used to combine multiple images of the same scene, each captured with a different focus setting, into a single all-in-focus image. By significantly improving image clarity and providing richer visual information, this technique has found wide application in fields such as microscopic imaging \cite{pei2021real}, visual sensor networks \cite{khan2022image}, and visual power patrol inspection \cite{chen2017application}.

\begin{figure}[t]
	\centering
	
	\includegraphics[width=.48\textwidth]{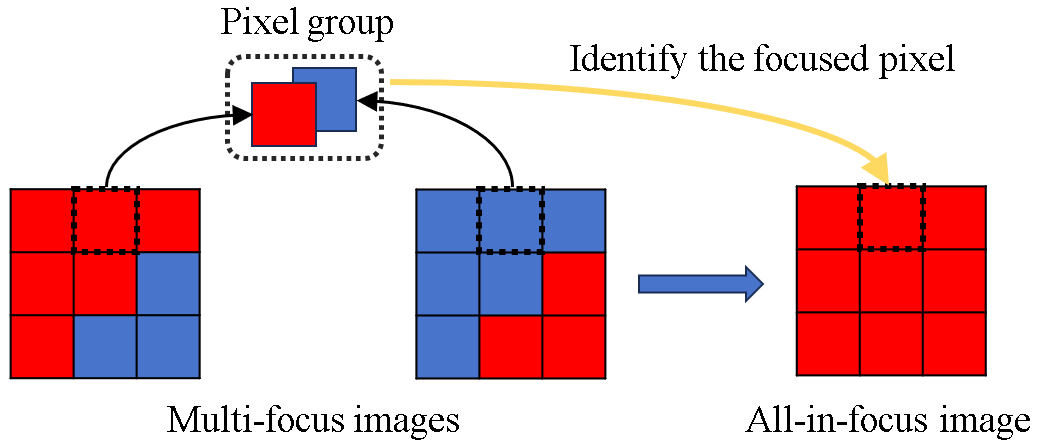}

	\caption{Cartoon schematic of multi-focus image fusion. Red boxes indicate focused pixels, while blue boxes represent defocused pixels.}
	\label{fig_0}
\end{figure}

Research on multi-focus image fusion algorithms has spanned several decades. Traditional methods typically operate in the spatial or transform domain to extract and integrate focus information from source images. These methods exhibit strong generalization capabilities; however, they rely on handcrafted features, making it difficult to estimate the focus level at each pixel accurately. Consequently, the fused images often suffer from reduced quality, such as blurred texture details or the presence of structural artifacts near the transition areas between focused and defocused regions. With the rise of deep learning, increasing attention has been directed toward deep learning-based multi-focus image fusion algorithms. These methods typically require large volumes of data to learn effective fusion strategies. Supervised approaches, for example, rely on precisely registered all-in-focus images as ground-truth, which are often difficult to obtain in practice. As a workaround, many studies \cite{tcsvt07_li2023} use synthetically generated data to train neural networks. However, synthetic data usually fails to fully replicate the complex focus distributions found in real images, which can lead to limited generalization when models are deployed in practical settings. To mitigate the need for labeled data,  unsupervised learning has become an active area of investigation. From a Bayesian perspective, these methods exploit image priors, such as gradient-based priors \cite{MFF-GAN} or deep image priors \cite{ZMFF}, to encourage the generation of all-in-focus outputs. Nonetheless, the priors employed in existing approaches are insufficient to precisely characterize the statistical and structural properties of focused regions, thereby impairing the accurate discrimination between focused and defocused pixels. Deep learning is promising for multi-focus image fusion due to its powerful feature extraction capabilities. However, current models are constrained by insufficiently realistic training data and suboptimal prior assumptions. As a result, their performance may fall short of expectations, and in some cases, may not offer significant improvements over traditional methods.

%This paper aims to address the limitations of existing deep learning-based multi-focus image fusion methods. From a pixel-level perspective, the fusion process can be reformulated as a classification task that distinguishes focused from defocused pixels. For each spatial location in the images, the corresponding pixels across all partially focused source images form a pixel group, where the number of pixels equals the number of input images (typically two). It is commonly assumed that only one pixel within each group is in focus. If the focused pixel can be correctly identified from each group and subsequently assembled, an all-in-focus image can be readily constructed. Motivated by this observation, we propose a novel deep learning framework for MFIF, termed Inter-image Pixel Shuffling (IPS). IPS's key idea is to independently perform focus classification for each pixel group formed across the source images. A major distinction between IPS and existing deep learning approaches is that IPS does not require any multi-focus images (either real or synthetic) for training. Instead, training only requires access to arbitrary natural optical images. Specifically, IPS treats a natural optical image and its low-pass filtered (blurred) version as the focused and defocused images. For each spatial position in the images, pixels between the two images are randomly swapped with a certain probability to form synthetic source images. A network is then trained to identify the focused pixel within each pixel group, using the original, unaltered optical image as the supervision signal. 

\begin{figure*}[t]
	\centering
	
	\includegraphics[width=.68\textwidth]{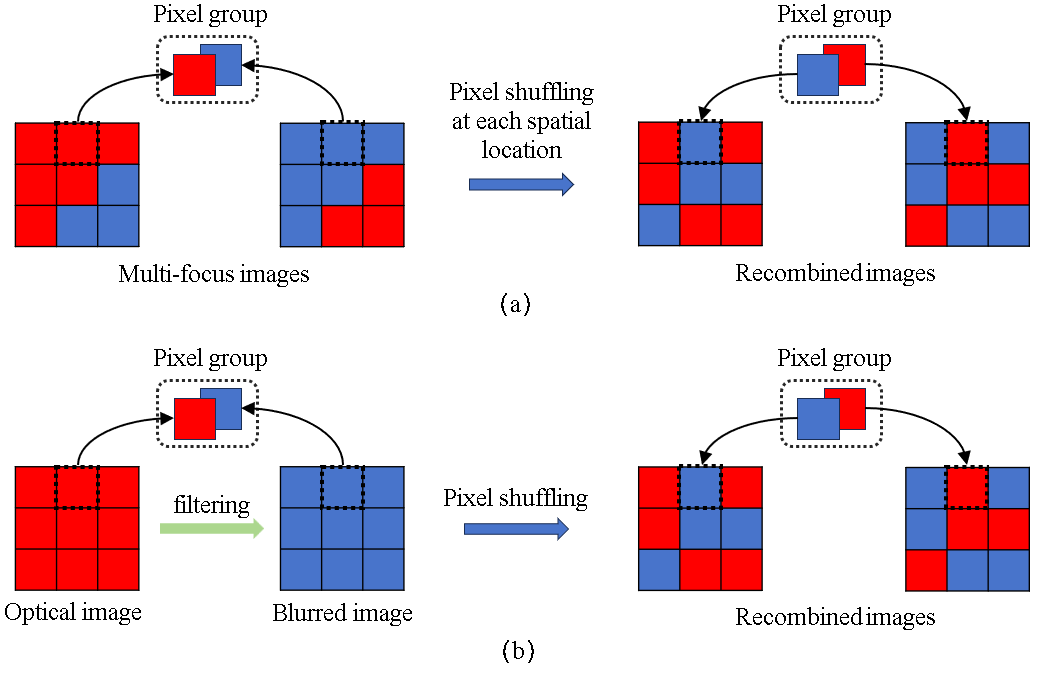}

	\caption{Illustration of pixel shuffling. Red boxes indicate focused pixels, while blue boxes represent defocused pixels. (a) Applying pixel shuffling to the multi-focus images yields a pair of recombined images. Since both the multi-focus and recombined image pairs share the same pixels, they correspond to the same all-in-focus image. (b) Filtering and pixel shuffling a single optical image can produce recombined images identical to those in (a). These recombined images can thus serve as substitutes for learning multi-focus image fusion, with the optical image in (b) used as the ground-truth.}

%\caption{Illustration of pixel shuffling. Red boxes indicate focused pixels, while blue boxes represent defocused pixels. By applying filtering and pixel shuffling, the multi-focus images (a) and an optical image (b) yield the same recombined images. The all-in-focus image corresponding to the recombined images in (a) and (b) is identical. Thus, we can use the recombined images in (b) to train multi-focus image fusion models and employ the optical image in (b) as the label.}
	
%	\caption{Illustration of pixel shuffling. Red boxes indicate focused pixels, while blue boxes represent defocused pixels. Exchanging pixels at the same spatial location across the source images does not alter the resulting all-in-focus image. By applying filtering and pixel shuffling, multi-focus images and an optical image may yield similar recombined images. In these recombined images, each pixel’s focus characteristic is independent of those of its neighboring pixels.}
	\label{motivation1}
\end{figure*} 

This paper aims to address the limitations of existing deep learning-based multi-focus image fusion methods. From a pixel-level perspective, MFIF can be formulated as a classification task to identify the focused pixel within each group of spatially aligned pixels from the source images, as shown in Fig. \ref{fig_0}. Each pixel group comprises pixels at the same spatial location across the source images (typically two), with the common assumption that only one pixel in the group is in focus. If the focused pixel can be correctly identified from each group and subsequently assembled, an all-in-focus image can be readily constructed. We further observe that shuffling the order of pixels within each pixel group, i.e., exchanging pixels at the same spatial coordinates across source images, does not alter the desired all-in-focus image, as shown in Fig. \ref{motivation1} (a). Based on the above observations, we propose a novel deep learning-based MFIF approach termed Inter-image Pixel Shuffling (IPS). The core idea is randomly swapping pixels at corresponding spatial locations across source images to construct training data for focused–defocused pixel classification. A notable advantage of IPS is that, due to its pixel-wise focus discrimination mechanism, it does not require multi-focus images (either real or synthetic data) for training. Instead, it can be trained using arbitrary optical images. Specifically, IPS treats a natural optical image and its low-pass filtered (blurred) version as the focused and defocused images. For each spatial position in the images, pixels between the two images are randomly swapped with a certain probability to form recombined source images, as shown in Fig. \ref{motivation1} (b). A network is then trained on the recombined images to learn to identify the focused pixel within each pixel group, using the original unfiltered optical image as the supervisory signal.
To ensure high-quality fusion, which requires both fine-grained spatial detail and holistic contextual understanding, IPS incorporates a cross-image fusion network architecture that synergistically combines convolutional neural networks (CNNs) for local feature representation and state space models for capturing global dependencies. This hybrid design enables the network to effectively reason about both local and non-local image structures.
%A network is then trained to identify the focused pixel within each pixel group, using the original, unfiltered optical image as the supervision signal. 
%On the other hand, reconstructing an all-in-focus image necessitates capturing the global contextual features present in the source images. To this end, IPS adopts a cross-image selective state space model as its backbone. This architecture effectively captures inter-image correlations and long-range dependencies among focused pixels by combining state space modeling with deep learning.
Extensive experiments conducted on multiple benchmark datasets demonstrate that IPS is capable of accurately extracting and fusing focus information, even in the absence of exposure to any multi-focus data during training. IPS substantially outperforms traditional fusion techniques and exhibits superior generalization ability compared to existing deep learning-based approaches.

%To further enhance the network's ability to capture long-range dependencies among focused regions, IPS incorporates a novel multi-scale selective state-space model as its backbone. This architecture integrates multi-scale feature extraction with sequential modeling, enabling the network to effectively distinguish focused pixels across a broader spatial context while maintaining computational efficiency. 

%Extensive experiments conducted on multiple benchmark datasets demonstrate that IPS is capable of accurately extracting and fusing focus information, even in the absence of exposure to any multi-focus data during training. IPS substantially outperforms traditional fusion techniques and exhibits superior generalization ability compared to existing deep learning-based approaches.

The primary contributions of this work are outlined as follows:

1. We propose IPS, a novel MFIF framework that reduces the reliance of deep learning-based approaches on real or synthetic multi-focus training data. IPS learns pixel-wise fusion rules using arbitrary single images, eliminating the need for dedicated multi-focus datasets. This approach offers greater flexibility and practicality than existing supervised and unsupervised methods, particularly in domains where acquiring large-scale multi-focus data is challenging, such as remote sensing and microscopic imaging.

%1. We propose IPS, a novel MFIF framework that reduces the reliance of deep learning-based approaches on real or synthetic multi-focus training data. IPS learns pixel-wise fusion rules using only natural optical images, thereby improving the applicability and generalizability of deep learning methods in practical scenarios, particularly in domains where acquiring paired multi-focus datasets is difficult, such as microscopic imaging and remote sensing.

2. We introduce an efficient cross-image fusion architecture that synergistically combines the localized feature extraction of CNNs with the global context modeling of state space models, allowing for robust identification of both local and non-local focus patterns across source images.

%2. We design a new multi-scale selective state-space network architecture, which effectively models long-range dependencies among focused pixels. By combining multi-scale representation learning with sequence modeling, this architecture enhances the network’s ability to discriminate between focused and defocused regions while maintaining computational efficiency.

\section{Related Work}
This section begins with a review of recent developments in multi-focus image fusion methods, which are generally classified into traditional and deep learning-based methods. We also summarize the advancements in the state space model.

\subsection{Traditional MFIF methods}
Traditional multi-focus image fusion methods can be broadly categorized into spatial domain methods and transform domain methods.

Spatial Domain Methods: These methods operate directly in the spatial domain by dividing images into different parts and evaluating their focus levels using metrics such as local variance and spatial frequency. A weight map is then generated for each source image to fuse the focused information. Depending on the image partitioning strategy, spatial domain methods are typically classified into block-based, region-based, and pixel-based techniques. Block-based methods \cite{Combination} segment the image into fixed-size blocks and apply fusion strategies, often involving a consistency test, on each block. These methods are sensitive to block size selection and may introduce grid-like artifacts in the fused image. Region-based methods \cite{LI2} assess focus levels over irregularly segmented regions, rather than fixed-size blocks, resulting in fused images that appear more natural than those produced by block-based methods. Pixel-based methods \cite{GuidedFiltering} perform fusion at the individual pixel level by evaluating the sharpness of each pixel and generating pixel-wise weight maps. While these approaches offer finer granularity and faster computation, they are more susceptible to noise, which can result in the presence of residual artifacts in the fused image.

Transform Domain Methods: These methods typically involve three stages: image transformation, coefficient fusion, and inverse transformation. Initially, the source images are transformed into a specific domain using a decomposition technique. The resulting coefficients are then fused based on predefined criteria, and finally, an inverse transform reconstructs the fused image. According to the adopted fusion strategies, transform domain methods are commonly divided into multi-scale decomposition, sparse representation, and gradient domain methods.
Multi-scale decomposition methods, such as pyramid \cite{sun2018multi} and wavelet \cite{bhatnagar2013directive} transforms, decompose the image into sub-bands at different scales, enabling feature fusion at each resolution level. 
Sparse representation-based methods \cite{ma2019multi} assume that the image can be represented as a linear combination of a small number of atoms from an overcomplete dictionary. Fusion is performed in the sparse coefficient space, followed by reconstruction using the learned dictionary. These methods are effective in noise suppression but may compromise the preservation of fine details, resulting in overly smoothed outputs. 
Gradient domain methods \cite{li2021multi} extract gradient features from the source images and reconstruct the final fused image using the fused gradients. This approach helps preserve fine details and structural information consistent with the source images. 

%In addition to the above, other representative transform domain fusion techniques include nonsubsampled contourlet transform, independent component analysis, compressed sensing, and discrete cosine transform, among others.

%Pyramid-based fusion performance is influenced by the decomposition depth, with deeper levels often resulting in blurred edges. Wavelet-based methods typically yield better fusion quality but require precise image registration to avoid performance degradation.

In summary, while traditional multi-focus image fusion algorithms have laid the groundwork for current research, they often struggle with issues such as insufficient detail preservation or the introduction of undesired artifacts, thereby limiting their effectiveness in practical scenarios.

\subsection{Deep learning-based MFIF methods}
With the growing success of deep learning across a broad range of computer vision tasks, it has also been widely adopted in the field of multi-focus image fusion. Depending on whether all-in-focus images are required as training labels, deep learning-based fusion methods can be classified into supervised and unsupervised approaches.

Supervised methods emphasize the design of effective neural network architectures to achieve higher fusion quality and computational efficiency. For example, Liu et al. \cite{liu2017_supervised} were among the first to introduce convolutional neural networks (CNNs) into multi-focus image fusion, employing a straightforward yet efficient CNN to learn a direct mapping from the source images to a focus map. Building upon this, Li et al. \cite{DRPL} developed DRPL, which utilizes a deep residual convolutional neural network (CNN) to produce a binary decision map directly from the source images. Structural similarity and edge-aware metrics are integrated into the training process to ensure that the resulting image is fully focused. To address fusion tasks involving more than two source images, Li et al. \cite{tcsvt07_li2023} proposed GRFusion, which employs a multidirectional gradient embedding strategy to detect hard pixels across multiple inputs, facilitating robust all-in-focus reconstruction. Recognizing that most fusion algorithms struggle near focus-defocus boundaries (FDBs), Ma et al. \cite{ma2020alpha} developed an $\alpha$-matte defocus boundary model for generating realistic synthetic data with defocus spread effects, and proposed a cascaded boundary-aware convolutional network, MMF-Net, which significantly improves fusion performance around FDB regions. Zhang et al. \cite{zhang2024exploit} designed a dual-branch network, DB-MFIF, which combines the strengths of end-to-end learning with decision map-based approaches to enhance fusion both near and far from FDBs. To improve model adaptability, Hu et al. \cite{hu2024incrementally} introduced the Incremental Network Prior Adaptation (INPA) framework, which leverages general knowledge from pretrained models and incorporates test-time priors to boost generalization. However, its two-stage optimization process increases computational cost and exhibits sensitivity to image misalignment. In addition, advanced architectures, including Transformers \cite{jin2023unsupervised}, generative adversarial networks (GANs) \cite{guo2019fusegan}, and convolutional neural network-based conditional random field models \cite{bouzos2023convolutional}, have also been investigated for enhancing multi-focus image fusion performance. Despite the promising results of supervised learning, these methods typically require large volumes of labeled data. As ground-truth all-in-focus images are difficult to obtain in practice, most approaches rely on synthetically generated datasets for training. However, the domain gap between synthetic and real-world data often hampers generalization performance in real-world scenarios.

Unsupervised MFIF methods have emerged to reduce reliance on labeled training data and enhance applicability to real-world scenes. In the absence of ground-truth, these methods leverage image priors, either explicit or implicit, to guide model learning. For example, based on the assumption that focused regions in source images should be preserved in the fused image, Zhang et al. \cite{MFF-GAN} proposed MFF-GAN, which encourages the generation of high-frequency textures through gradient and intensity contrast constraints. Similarly, Ma et al. \cite{Sesf-fuse} introduced SESF-Fuse, which utilizes the gradient of deep features to construct decision maps, relying on a high-frequency consistency prior. However, since gradients are less responsive in areas with weak texture, such approaches may lead to suboptimal fusion. To address this limitation, Liu et al. \cite{liu2023focus} proposed a network that integrates gradient and local variation to better capture texture in flat regions. Another line of work exploits the concept of deep image priors, which refers to the intrinsic tendency of CNNs to capture low-frequency features. Hu et al. \cite{ZMFF} combined this prior with a deep mask prior to propose ZMFF, a zero-shot multi-focus fusion method. Nonetheless, it remains sensitive to image misalignment and defocus spread effects. More recently, proxy-task-driven methods have shown potential in implicitly embedding structural priors. Lin et al. \cite{lin2024fusion2void} formulated image inpainting as a proxy task for multi-focus image fusion and proposed Fusion2Void, where the fused output guides the reconstruction of high-frequency details in partially focused images, only achievable when the fusion is truly all-in-focus. Similarly, Quan et al. \cite{quan2024dual} proposed UDPNet, which treats denoising as a proxy task. By applying a blind-spot self-supervised denoising framework to estimate the fusion mask, UDPNet reconstructs the final all-in-focus image after denoising.

Overall, while deep learning offers significant potential for multi-focus image fusion due to its strong representation capacity, the lack of large-scale labeled datasets limits the scalability of supervised approaches. Unsupervised methods offer a more flexible alternative, yet the priors currently used, such as gradient consistency, are often insufficient to accurately model focus characteristics, leaving room for further improvement in fusion performance.
\begin{figure*}[ht]
	\centering
	\includegraphics[width=1\textwidth]
	{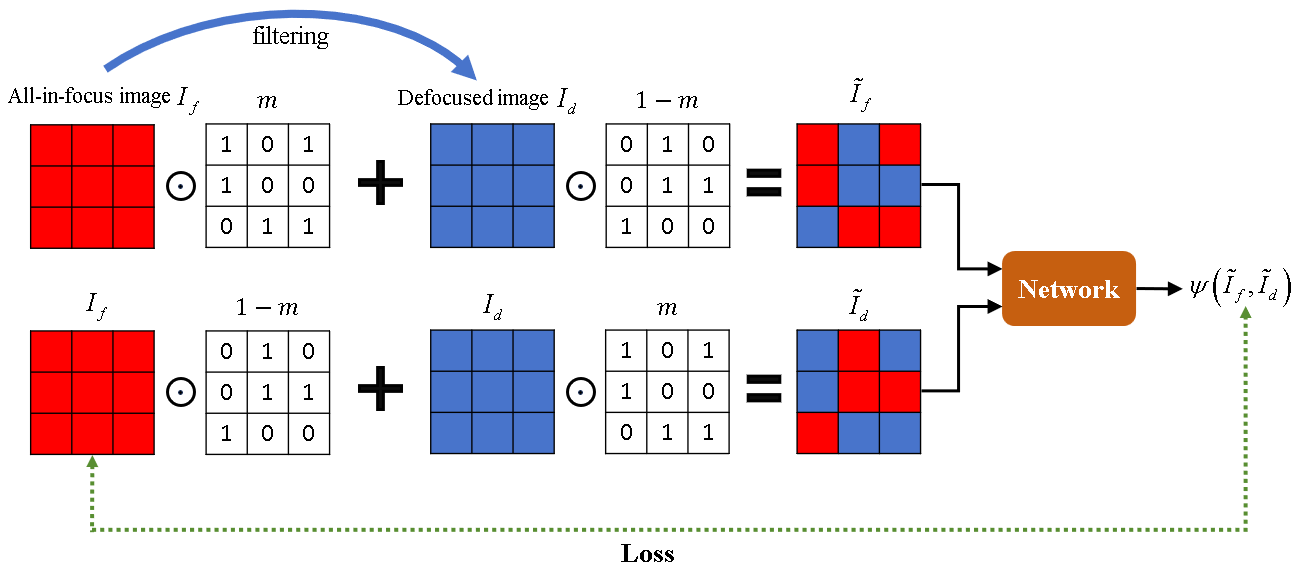}
	\caption{\label{FusionNet} IPS training pipeline. Red boxes indicate focused pixels, while blue boxes represent defocused pixels. The mask $m$ shares the same dimensions as the image.}
\end{figure*}

\subsection{State space models}

Convolutional Neural Networks (CNNs) and Transformers are the predominant backbone architectures in image fusion tasks \cite{zhou2024general}, each exhibiting distinct strengths and limitations. CNNs are proficient in extracting local features due to their localized receptive fields and benefit from parallelizable convolution operations, resulting in linear computational complexity concerning input size. This efficiency makes them particularly suitable for deployment on resource-constrained edge devices. However, their limited capacity to capture long-range dependencies hinders their performance in scenarios requiring holistic image understanding. In contrast, Transformers possess a strong capability for modeling global context. Nonetheless, their computational complexity scales quadratically with input resolution, significantly burdening memory and processing resources.

To overcome the limitations of CNNs and Transformers, structured state space models (SSMs) have recently garnered attention for their ability to model long-range dependencies with linear computational complexity. The S4 \cite{gu2021efficiently} model introduced a structured formulation of deep SSMs, laying the groundwork for their practical application. Building on S4, the S5 \cite{smith2022simplified} model introduced MIMO SSM and efficient parallel scanning. The H3 \cite{fu2022hungry} architecture subsequently advanced the performance of SSMs in natural language processing (NLP), narrowing the accuracy gap with Transformer-based models. More recently, Mamba \cite{gu2023mamba}, a data-dependent SSM incorporating selective scanning, has demonstrated superior performance over Transformers in NLP tasks, highlighting its capability to balance global dependency modeling with computational efficiency. Encouraged by these advances, SSMs have begun to find application in various vision tasks, including image classification \cite{liu2024vmamba}, restoration \cite{guo2024mambair}, and segmentation \cite{xing2024segmamba}. In the field of image fusion, several pioneering SSM-based models such as FusionMamba \cite{xie2024fusionmamba} and Pan-Mamba \cite{he2025pan} have emerged, significantly advancing the state-of-the-art. However, in the specific domain of multi-focus image fusion, the design of effective SSM architectures remains largely underexplored, presenting a valuable avenue for future investigation.

%%%%%%%%%%%%%%%%%%%%%%%%%%%%%

%\begin{figure*}[ht]
%\centering
%\includegraphics[width=0.8\textwidth]
%{framework.png}
%\caption{\label{FusionNet} IPS training pipeline. Red boxes indicate focused pixels, while blue boxes represent defocused pixels. The mask m shares the same dimensions as the image.}
%\end{figure*}

\section{Proposed Method}

The overall training pipeline of the proposed IPS framework is illustrated in Fig. \ref{FusionNet}. During training, IPS leverages arbitrary optical images without requiring them to be all-in-focus, along with their corresponding low-pass filtered counterparts. At inference time, the trained network can be directly applied to real multi-focus images. In this study, we demonstrate IPS using two source images. However, the framework can be readily extended to handle more than two source images. 

\begin{figure*}[ht]
	\centering
	\includegraphics[width=1\textwidth]
	{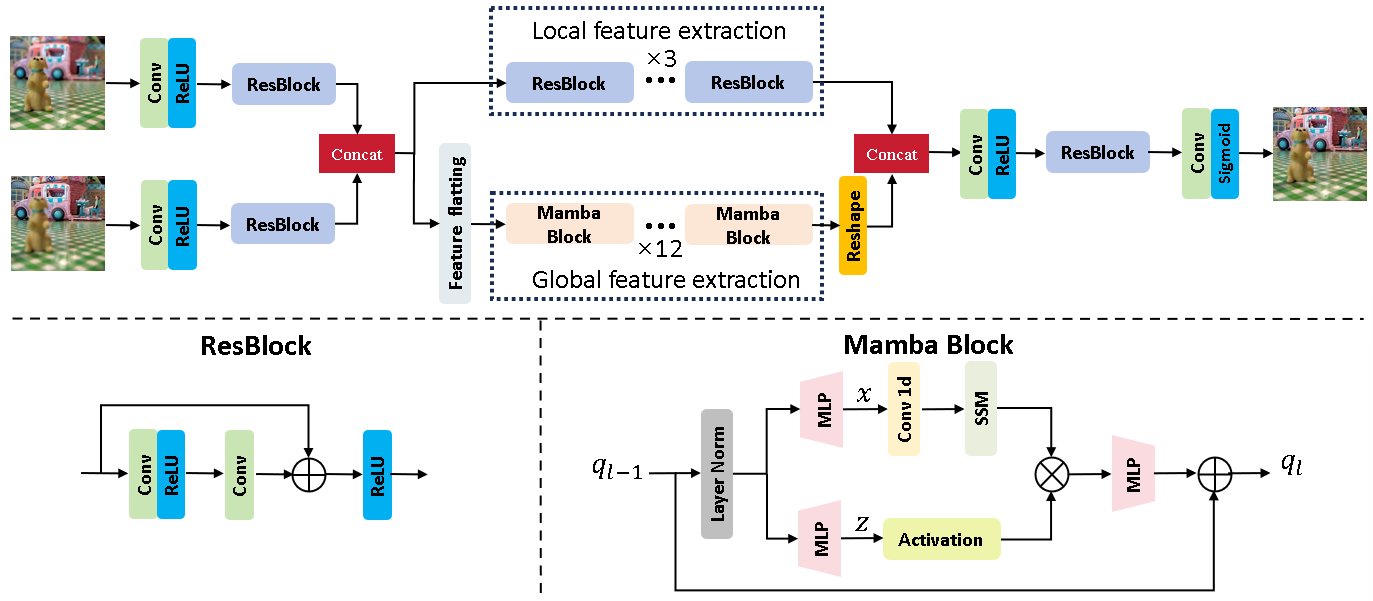}
	\caption{\label{Network} Architecture of the Cross-Image Fusion Network.}
\end{figure*}

\subsection{Inter-image pixel shuffling}
IPS formulates multi-focus image fusion as a pixel-wise classification problem. Given an optical image $I_f \in \mathbb R^{H \times W \times J}$, where $H$, $W$, and $J$ denote the height, width, and number of channels, respectively, $I_f$ is regarded as an all-in-focus image. A blurred version $I_d$ is generated by applying a low-pass filter to $I_f$, resulting in a pair of spatially aligned images. Pixels in $I_f$ are considered focused, while those in $I_d$ are assumed to be defocused. At each spatial location, the corresponding pixels from $I_f$ and $I_d$ constitute a pixel group,
\begin{align}
	\label{eq:PG}
	\mathcal{P}_{g}^{(h,w,j)} = \left\{ {I_f^{(h,w,j)},I_d^{(h,w,j)}} \right\},
\end{align}
where $I_f^{(h,w,j)}$ denotes the pixel at location $(h,w,j)$ in image $I_f$, $I_d^{(h,w,j)}$ is defined analogously. Each group contains as many pixels as there are source images (in this case, two), assuming that exactly one pixel in each group is in focus.

The objective of IPS is to train a neural network to classify which pixel within each group is focused. However, if the focused and defocused pixels are always arranged in a fixed order across all groups, the network may learn to rely on positional bias rather than focus-related features. This can lead to overfitting and limit the model’s ability to generalize. To address this, IPS introduces randomization by shuffling the order of pixels within each group with a predefined probability $p$. This stochastic permutation ensures that the spatial arrangement of focused pixels varies across training samples. Consequently, the network must infer the focus level of each pixel independently to accurately identify the focused pixel in each group.

IPS achieves pixel shuffling across all pixel groups simultaneously by applying a randomly generated binary mask. Specifically, the images $I_f$ and $I_d$ are masked and recombined as 
%\begin{align}
%	\label{eq:2}
%	{{\tilde I}_f} &= {I_f} \odot m + {I_d} \odot \left( {1 - m} \right)\\
%		\label{eq:32}
%	{{\tilde I}_d} &= {I_f} \odot \left( {1 - m} \right) + {I_d} \odot m
%\end{align}
\begin{equation}
	\label{eq:2}
	%	\scriptsize
	%	\left
	% MathType!Translator!2!1!LaTeX.tdl!LaTeX 2.09 and later!
	\begin{aligned}
	{{\tilde I}_f} &= {I_f} \odot m + {I_d} \odot \left( {1 - m} \right)\\
{{\tilde I}_d} &= {I_f} \odot \left( {1 - m} \right) + {I_d} \odot m
	\end{aligned}
\end{equation}
where $m \in {\left\{ {0,1} \right\}^{H \times W \times J}}$ is a randomly sampled binary mask. 
Each element in $m$ is independently set to 0 with probability $p$, and to $1$ with probability $1 - p$. 
A new mask $m$ is generated at each training iteration. According to Eq. (\ref{eq:2}), this masking strategy results in a random exchange of pixels between $I_f$ and $I_d$ at each spatial location. The resulting image ${{\tilde I}_f}$ is no longer entirely in focus like the original $I_f$, and ${{\tilde I}_d}$ is no longer entirely defocused like $I_d$. Instead, both ${{\tilde I}_f}$ and ${{\tilde I}_d}$ contain a mixture of focused and defocused pixels, resembling multi-focus images. These two mixed-focus images, ${{\tilde I}_f}$ and ${{\tilde I}_d}$, are then used as input to the network, which is trained to estimate per-pixel focus levels. The corresponding loss function can be expressed as 
\begin{equation}
	\label{eq:loss}
	\begin{aligned}
	{\mathcal{L}_{fuse}} = {\left\| {\psi \left( {{{\tilde I}_f},{{\tilde I}_d}} \right) - {I_f}} \right\|_1}
	\end{aligned}
\end{equation}
where $\psi$ represents the fusion network used in IPS. As defined by the loss function in Eq. (\ref{eq:loss}), the network learns to identify the focused pixel from each pixel group at every spatial location within the input images. This fine-grained, pixel-level focus discrimination ability allows IPS to generalize effectively to real multi-focus images, despite the absence of such data during training. Moreover,  during testing, IPS can directly process multi-focus images without the need for additional masking, owing to its learned capacity for pixel-wise focus-defocus classification.

\subsection{Cross-image fusion network}
The architecture of the cross-image fusion network adopted by IPS is illustrated in Fig. \ref{Network}. Each source image is first processed through a convolutional layer, followed by a ResBlock to extract shallow features. These features are then concatenated along the channel dimension and fed into two parallel branches: one comprising 3 ResBlocks and the other containing 12 Mamba blocks. The ResBlock branch is designed to extract local image features, enhancing fine structural details in the fused image. In contrast, the Mamba block captures global contextual dependencies, facilitating the identification of spatially distant but semantically related focused pixels across the source images. The detailed configurations of both the ResBlock and Mamba block are shown in Fig. \ref{Network}.

As introduced in related works \cite{gu2023mamba, he2025pan}, Mamba is a data-dependent state space model. A classical state space model can be formulated as a linear time-invariant (LTI) system, which maps a one-dimensional input sequence $x(t) \in \mathbb R$ to an output sequence $y(t) \in \mathbb R$ via a hidden state $s(t) \in \mathbb R^N$. Such a system can be described by an ordinary differential equation (ODE) of the form,
\begin{equation}
	\label{eq:ODE}
	\begin{aligned}
		s'\left( t \right) &= As\left( t \right) + Bx\left( t \right)\\
	y\left( t \right) &= Cs\left( t \right) + Dx\left( t \right)
	\end{aligned}
\end{equation}
where $N$ is the state size, $A \in \mathbb R^{N \times N}$, $B \in \mathbb R^{N \times 1}$, $C \in \mathbb R^{1 \times N}$ and $D \in \mathbb R$.

To enable implementation within deep learning frameworks, this continuous formulation is typically discretized. Specifically, a step size parameter $\Delta$ is introduced to convert the continuous parameters $A$ and $B$ into their discrete counterparts $\bar A$ and $\bar B$. A commonly used discretization technique is the Zero-Order Hold (ZOH) method, defined as
\begin{equation}
	\label{eq:6}
	\begin{aligned}
			\bar A &= \exp \left( {\Delta A} \right),\\
		\bar B &= {\left( {\Delta A} \right)^{ - 1}}\left( {\exp \left( A \right) - I} \right) \cdot \Delta B.
	\end{aligned}
\end{equation}

This discretized system can be rewritten as the following RNN form,
\begin{equation}
	\label{eq:7}
	\begin{aligned}
		{s_k}& = \bar A{s_{k - 1}} + \bar B{x_k},\\
		{y_k}& = C{s_k} + D{x_k}.
	\end{aligned}
\end{equation}

Furthermore, the output of Eq. (\ref{eq:7}) is equivalent to performing a global convolution on the input
\begin{equation}
	\label{eq:8}
	\begin{aligned}
			\bar K &= \left( {C\bar B,C\bar A\bar B,...,C{{\bar A}^{L - 1}}\bar B} \right),\\
		y &= x*\bar K.
	\end{aligned}
\end{equation}
where $L$ denotes the length of the input sequence, $\bar K$ is a structured convolutional kernel, $*$ represents the convolution operation. 

To accommodate the two-dimensional nature of images, IPS first flattens the extracted 2D feature maps into one-dimensional token sequences before feeding them into the Mamba blocks. These blocks are designed to capture global contextual information and model long-range dependencies across spatial locations. Specifically, the input sequence $q_{l-1}$ is first processed with layer normalization, followed by two parallel multilayer perceptrons (MLPs) that project it into two intermediate sequences, $x$ and $z$. The sequence $x$ is processed by a 1D convolutional layer combined with a SiLU activation function to produce $x'$, which is used for estimating data-dependent parameters for SSM. The processed sequence $x'$ is projected onto the parameters $A$ and $B$, and a step size parameter $\Delta$ is introduced to discretize these parameters into their discrete counterparts $\bar{A}$ and $\bar{B}$. The resulting features from the 1D convolution are then passed through an SSM, whose output is gated by $z$. The resulting representation is then passed through an additional MLP and added to the original input $q_{l-1}$, producing the final output sequence $q_{l}$. The SSM operation is describe in Eq. (\ref{eq:8}). In the global feature extraction branch, IPS employs a stack of 12 Mamba blocks. The output of the final blcok is reshaped back into a 2D feature map and concatenated with the local features extracted via the parallel ResBlock branch. This fused feature is subsequently used to reconstruct the final all-in-focus image.

\subsection{Parameter setting}

IPS can be trained using arbitrary optical images without requiring them to be all-in-focus. This work uses 4,744 natural images from the Waterloo Exploration Database \cite{ma2016waterloo} as the training dataset. In each training iteration, a mean filter with a kernel size randomly selected from the range [3, 31] is applied to a source image $I_f$ to produce a blurred counterpart $I_d$. Subsequently, a stochastic binary mask $m$ is introduced to generate input samples for the pixel classification task, as defined in Eq. (\ref{eq:2}). Each element in $m$ is independently set to zero with a probability $p=0.5$. Furthermore, data augmentation is performed by randomly swapping the order of the source images during training.

%Each element in $m$ is independently set to zero with a probability $p$, where $p$ is dynamically sampled from a uniform distribution over the interval [0, 1] at every iteration.

% To construct the input source images, IPS employs a random binary mask $m$, as described in Eq. (\ref{eq:2}). Each element in the mask $m$ is set to zero with a probability $p$. The value of $p$ varies dynamically during training and is randomly sampled from the range [0, 1] at each iteration.
  The model is implemented using the PyTorch framework and trained on a single NVIDIA GeForce GTX 3080Ti GPU. Training is conducted for 1,000,000 iterations with a batch size of 1. Network weights are updated using the Adam optimizer, with an initial learning rate of 0.0001. The learning rate remains constant for the first 500,000 iterations and then linearly decays to zero over the remaining 500,000 iterations.

\begin{table}[!t]
\centering
	\small
	\renewcommand{\tabcolsep}{1.5pt} 
        \renewcommand{\arraystretch}{1.7}
        \captionsetup{
          singlelinecheck=off, 
          justification=centering, 
          labelsep=newline, 
          textformat=simple, 
        }
         \caption{
          {Quantitative results on the Lytro dataset. Best results are shown in \textcolor{red}{red} and the second-best in \textcolor{blue}{blue}.}}
	\begin{tabular}{c|cccccc}
		\hline
		Methods & \(\mathcal{Q}_{MI}\) & \(\mathcal{Q}_{SF}\) & \(\mathcal{Q}_{S}\) & \(\mathcal{Q}_{CB}\) & \(\mathcal{Q}_{AB/F}\) & \(\mathcal{Q}_{NCIE}\) \\
		\hline

  		DSIFT \cite{DSIFT}  &1.0271 & 0.0685 & 0.8158 & 0.7200 & 0.6760 & 0.8194\\ 
		DCT\_Corr \cite{DCT_Corr}  &1.0197 & 0.0674 & 0.8173 & 0.7103 & 0.6419 & 0.8191\\
        
            \hline
            
		CNN \cite{liu2017_supervised}  &1.0124& 0.0748 & 0.8373 & \textcolor{blue}{0.7878} & 0.7260 & 0.8325\\
		IFCNN \cite{IFCNN}  &1.0325& 0.0745 & 0.8253 & 0.7169 & 0.7206 & 0.8270\\
		SwinFusion \cite{Swinfusion}  &1.0217& 0.0735 & 0.8302 & 0.7218 & 0.7140 & 0.8237\\
         
            \hline

%		GCF \cite{GCF}  &1.0786& 0.0743 & 0.8370 & 0.7756 & 0.7240 & 0.8292\\
%		FusionDN \cite{FusionDN}  &0.7151& 0.0711 & 0.7614 & 0.5703 & 0.6000 & 0.8199\\
		U2Fusion \cite{U2fusion}  &0.7027& 0.0721 & 0.7213 & 0.5533 & 0.6089 & 0.8191\\
		MFF-GAN \cite{MFF-GAN}  &0.7584& 0.0723 & 0.8067 & 0.6253 & 0.6567 & 0.8215\\
%  		{DRPL} \cite{DRPL}  &{1.0178}& {0.0746} & {0.8480} & {0.7760} & {0.7421} & {0.8343} \\
%  		{MGIM} \cite{MGIM}  &{1.0124}& {0.0588} & {0.7496} & {0.7251} & {0.6282} & {0.8337} \\
		ZMFF \cite{ZMFF}  &0.8466& 0.0752 & 0.8407 & 0.7165 & 0.6996 & 0.8253\\
		Fusion2Void \cite{lin2024fusion2void}&\textcolor{red}{1.0842} &\textcolor{blue}{0.0753} & \textcolor{blue}{0.8482} & \textcolor{red}{0.7895}  &{0.7424}  &{0.8391} \\
        \hline
		
		\textbf{Ours}&\textcolor{blue}{1.0572} &\textcolor{red}{0.0769} & \textcolor{red}{0.8618} & {0.7692}  &\textcolor{red}{0.7461}  &\textcolor{red}{0.8392} \\
		\hline
	\end{tabular}%
	\label{t1}%
\end{table}

\begin{table}[!t]
\centering
	\small
	\renewcommand{\tabcolsep}{1.5pt} 
        \renewcommand{\arraystretch}{1.7}
        \captionsetup{
          singlelinecheck=off, 
          justification=centering, 
          labelsep=newline, 
          textformat=simple, 
        }
	\caption{
  {Quantitative results on the MFFW dataset. Best results are shown in \textcolor{red}{red} and the second-best in \textcolor{blue}{blue}.}}
	\begin{tabular}{c|cccccc}
		\hline
		Methods & \(\mathcal{Q}_{MI}\) & \(\mathcal{Q}_{SF}\) & \(\mathcal{Q}_{S}\) & \(\mathcal{Q}_{CB}\) & \(\mathcal{Q}_{AB/F}\) & \(\mathcal{Q}_{NCIE}\) \\
		\hline

  		DSIFT \cite{DSIFT}  &0.6902 & 0.0835 & 0.6460 & 0.6210 & 0.6064 & 0.8168\\
		DCT\_Corr \cite{DCT_Corr}  &0.6897 & 0.0838 & 0.6547 & 0.6087 & 0.5958 & 0.8168\\
        
            \hline
            
		CNN \cite{liu2017_supervised}  &0.7152& 0.0833 & 0.6423 & 0.6207 & 0.6048 & 0.8175\\
		IFCNN \cite{IFCNN}  &0.7003& 0.0853 & 0.6635 & 0.5882 & 0.5910 & 0.8170\\
		SwinFusion \cite{Swinfusion}  &0.7001& 0.0621 & 0.6333 & 0.5650 & 0.5834 & 0.8167\\
         
            \hline

%		GCF \cite{GCF}  &0.7171& 0.0864 & 0.6415 & 0.6181 & 0.6069 & 0.8176\\
%		FusionDN \cite{FusionDN}  &0.6543& 0.0802 & 0.5653 & 0.4686 & 0.4889 & 0.8158\\
		U2Fusion \cite{U2fusion}  &0.6436& 0.0879 & 0.5747 & 0.5183 & 0.5281 & 0.8148\\
		MFF-GAN \cite{MFF-GAN}  &0.6617& \textcolor{blue}{0.0919} & 0.6255 & 0.5491 & 0.5591 & 0.8158\\
%  		{DRPL} \cite{DRPL}  &{0.7781}& {0.0869} & {0.7291} & {0.6457} & {0.6753} & {0.8214} \\
%  		{MGIM} \cite{MGIM}  &{0.7762}& {0.0653} & {0.6490} & {0.6379} & {0.5983} & {0.8170} \\
		ZMFF \cite{ZMFF}  &0.7274& 0.0858 & 0.7179 & 0.6326 & 0.6509 & 0.8182\\
			Fusion2Void \cite{lin2024fusion2void}	&\textcolor{blue}{0.7881} &\textcolor{red}{0.0920} & \textcolor{blue}{0.7294} & \textcolor{blue}{0.6457}  &\textcolor{blue}{0.6758}  &\textcolor{blue}{0.8216} \\
        \hline
		
		\textbf{Ours}&\textcolor{red}{0.9409} &{0.0880} & \textcolor{red}{0.7395} & \textcolor{red}{0.6710}  &\textcolor{red}{0.7100}  &\textcolor{red}{0.8271} \\
		\hline
	\end{tabular}%
	\label{t2}%
\end{table}

%\begin{table}[]
%	\centering
%	\small
%	\renewcommand{\tabcolsep}{1.5pt} 
%	\renewcommand{\arraystretch}{1.7}
%	\captionsetup{
%		singlelinecheck=off, 
%		justification=centering, 
%		labelsep=newline, 
%		textformat=simple,
%	}
%	\caption{
%		{Quantitative results on the Real-MFF and MFI-WHU datasets. Best results are shown in \textcolor{red}{red} and the second-best in \textcolor{blue}{blue}.}}
%	% 使用 tabularx 宽度自适应栏宽
%	\begin{tabularx}{\linewidth}{c|*{2}{>{\centering\arraybackslash}X}|*{2}{>{\centering\arraybackslash}X}}
%		\hline
%		\multirow{2}{*}{Methods} & \multicolumn{2}{c|}{Real-MFF} & \multicolumn{2}{c}{MFI-WHU} \\ \cline{2-5}
%		& SSIM & PSNR & SSIM & PSNR \\ \hline
%		DSIFT \cite{DSIFT}         & 0.987 & 39.94 & 0.992 & \textcolor{blue}{45.26} \\
%		DCT\_corr \cite{DCT_Corr}    & 0.982 & 30.65 & 0.990 & 42.97 \\ \hline
%		CNN \cite{liu2017_supervised}          & 0.988 & 40.55 & 0.993 & 45.05 \\ 
%		IFCNN \cite{IFCNN}        & \textcolor{blue}{0.990} & \textcolor{blue}{40.77} & 0.993 & 40.61 \\ 
%		SwinFusion \cite{Swinfusion}   & 0.978 & 28.09 & 0.951 & 31.51 \\ \hline
%		U2Fusion \cite{U2fusion}     & 0.958 & 31.70 & 0.871 & 24.23 \\
%		MFF-GAN \cite{MFF-GAN}      & 0.963 & 30.54 & 0.967 & 30.49 \\ 
%		ZMFF  \cite{ZMFF}        & 0.958 & 33.18 & 0.966 & 32.13 \\ 
%		Fusion2Void \cite{lin2024fusion2void}  & 0.966 & 37.02 & \textcolor{blue}{0.996} & 44.43 \\ \hline
%		\textbf{Ours} & \textcolor{red}{0.991} & \textcolor{red}{42.19} & \textcolor{red}{0.997} & \textcolor{red}{47.52} \\ \hline
%	\end{tabularx}
%		\label{ssimpsnr}
%\end{table}

\begin{table}[]
	\centering
	\small
	\renewcommand{\tabcolsep}{1.5pt} 
	\renewcommand{\arraystretch}{1.7}
	\captionsetup{
		singlelinecheck=off, 
		justification=centering, 
		labelsep=newline, 
		textformat=simple,
	}
	\caption{
		{Quantitative results on the Real-MFF and MFI-WHU datasets. Best results are shown in \textcolor{red}{red} and the second-best in \textcolor{blue}{blue}.}}
	% 使用 tabularx 宽度自适应栏宽
	\begin{tabularx}{\linewidth}{c|*{2}{>{\centering\arraybackslash}X}|*{2}{>{\centering\arraybackslash}X}}
		\hline
		\multirow{2}{*}{Methods} & \multicolumn{2}{c|}{Real-MFF} & \multicolumn{2}{c}{MFI-WHU} \\ \cline{2-5}
		& PSNR& SSIM & PSNR & SSIM  \\ \hline
		DSIFT \cite{DSIFT}        & 39.94 & 0.987  & \textcolor{blue}{45.26}& 0.992  \\
		DCT\_corr \cite{DCT_Corr}   & 30.65 & 0.982 & 42.97 & 0.990  \\ \hline
		CNN \cite{liu2017_supervised}        & 40.55  & 0.988 & 45.05 & 0.993  \\ 
		IFCNN \cite{IFCNN}       & \textcolor{blue}{40.77}  & \textcolor{blue}{0.990} & 40.61 & 0.993  \\ 
		SwinFusion \cite{Swinfusion} & 28.09  & 0.978 & 31.51 & 0.951  \\ \hline
		U2Fusion \cite{U2fusion}    & 31.70 & 0.958 & 24.23 & 0.871  \\
		MFF-GAN \cite{MFF-GAN}     & 30.54 & 0.963 & 30.49 & 0.967  \\ 
		ZMFF  \cite{ZMFF}      & 33.18  & 0.958 & 32.13 & 0.966  \\ 
		Fusion2Void \cite{lin2024fusion2void} & 37.02 & 0.966  & 44.43  & \textcolor{blue}{0.996} \\ \hline
		\textbf{Ours} & \textcolor{red}{42.19} & \textcolor{red}{0.991} & \textcolor{red}{47.52} & \textcolor{red}{0.997}  \\ \hline
	\end{tabularx}
	\label{ssimpsnr}
\end{table}

		\begin{figure*}[t]
	\centering
	\includegraphics[width=\textwidth]{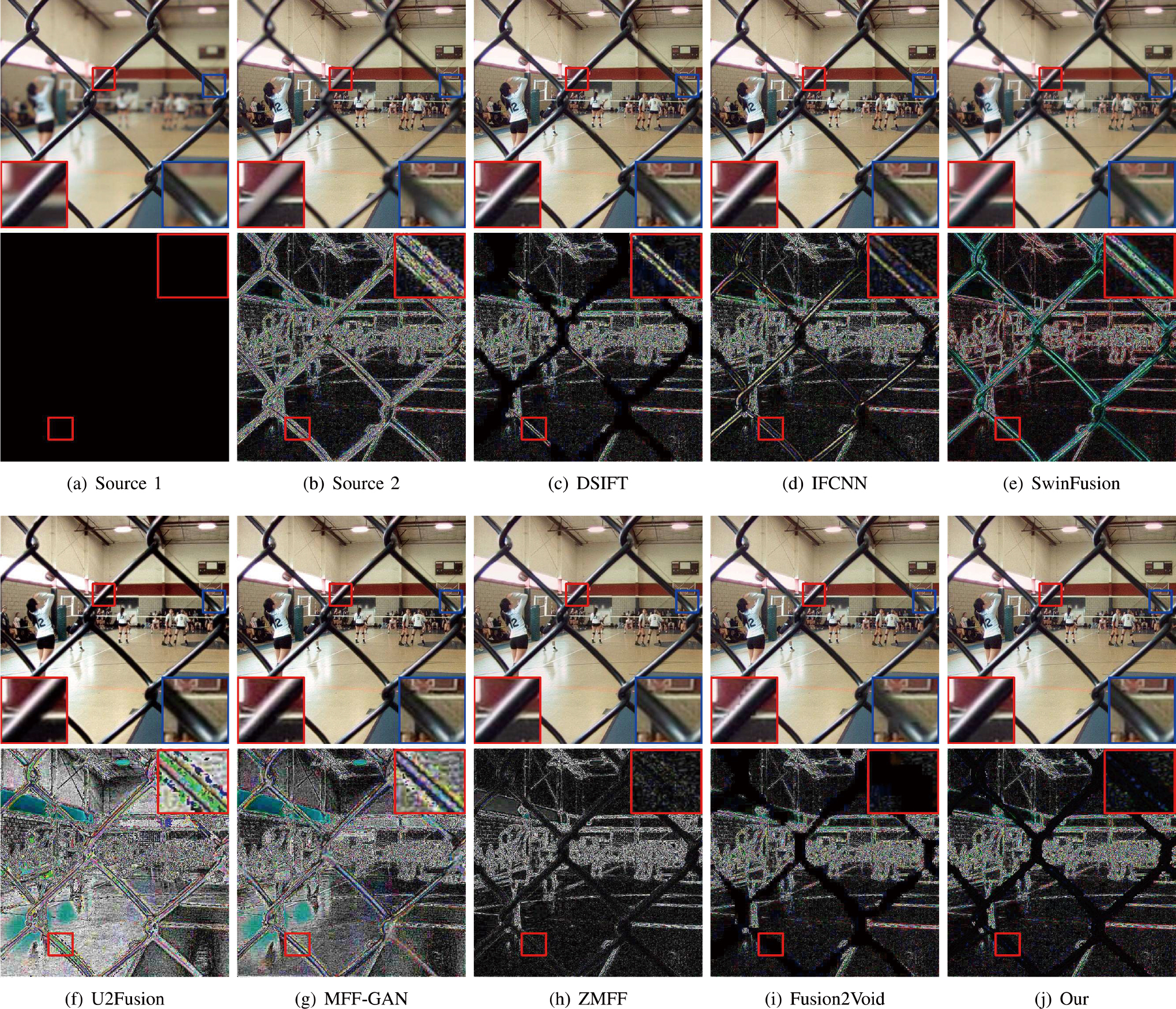}
	\caption{Visual comparison of fused images on the Lytro dataset. Corresponding difference maps with respect to Source Image 1 are also presented.}
	\label{fig_lytro}
\end{figure*}

		\begin{figure*}[t]
	\centering
	\includegraphics[width=\textwidth]{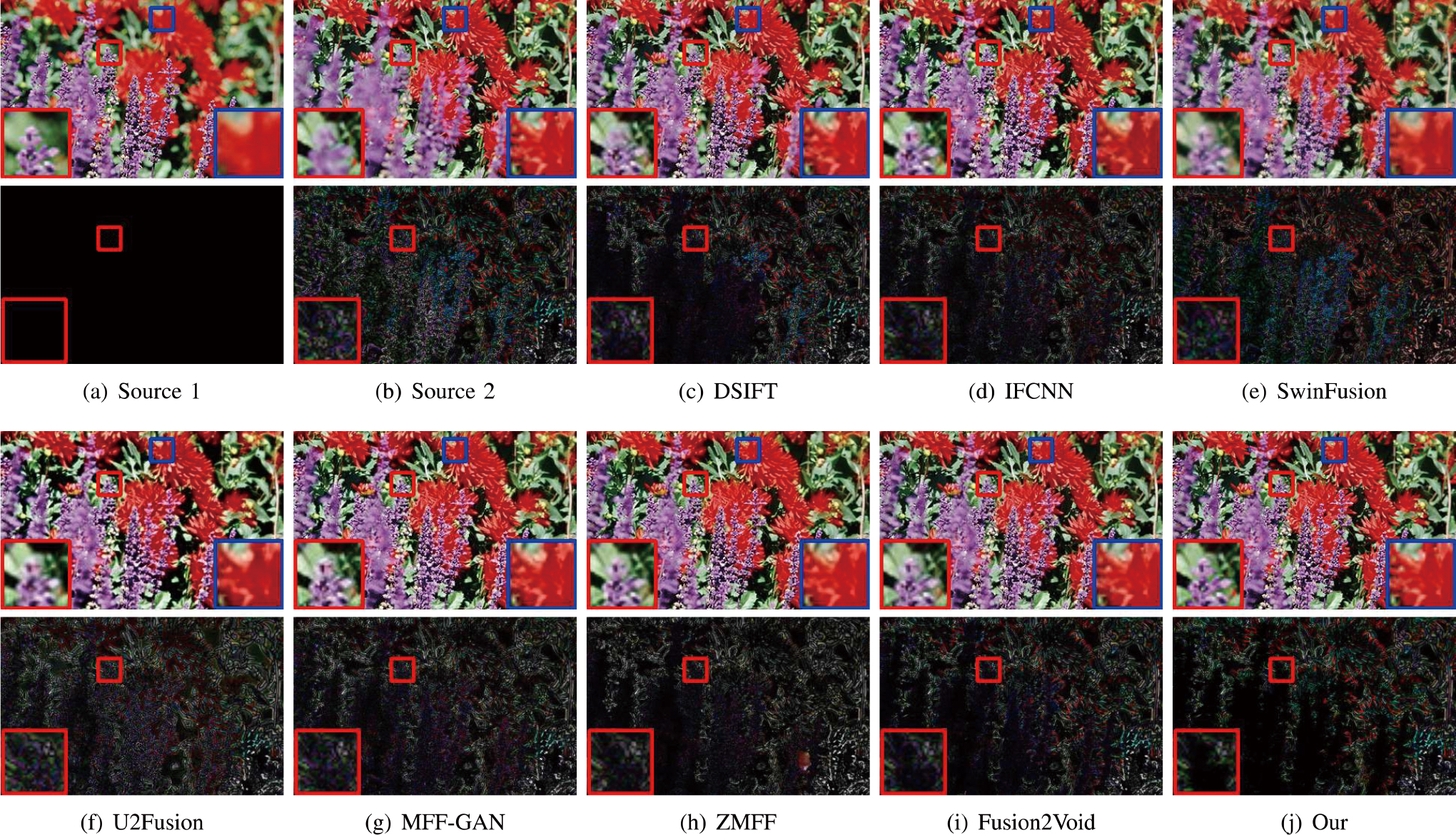}
	\caption{Visual comparison of fused images on the MFFW dataset. Corresponding difference maps with respect to Source Image 1 are also presented.}
	\label{fig_MFFW}
\end{figure*} 

		\begin{figure*}[t]
	\centering
	\includegraphics[width=\textwidth]{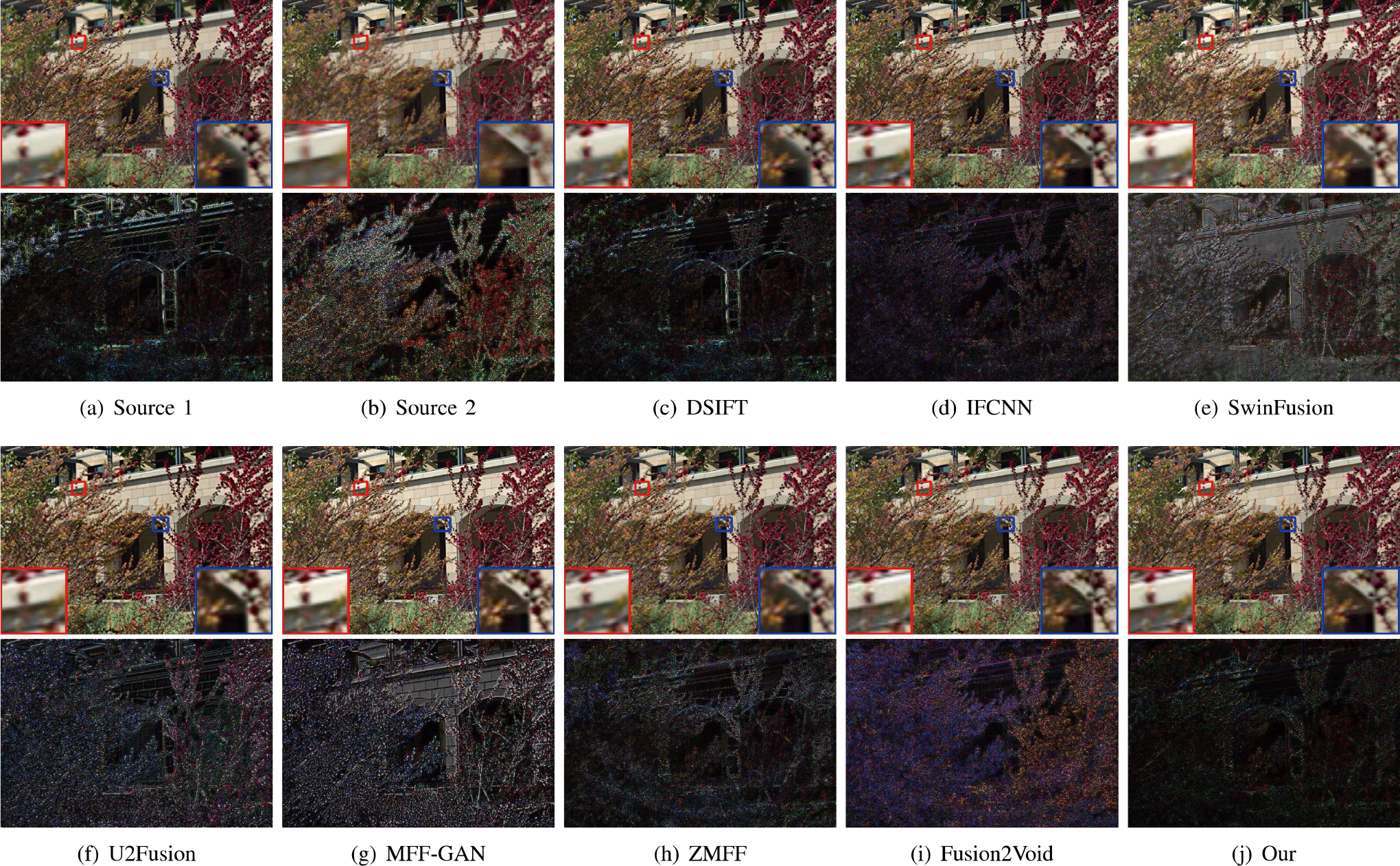}
	\caption{Visual comparison of fusion results on the Real-MFF dataset. Difference maps with respect to the ground-truth all-in-focus images are provided for reference.}
	\label{fig_realMFF}
\end{figure*}

\begin{figure*}[t]
\centering
\includegraphics[width=\textwidth]{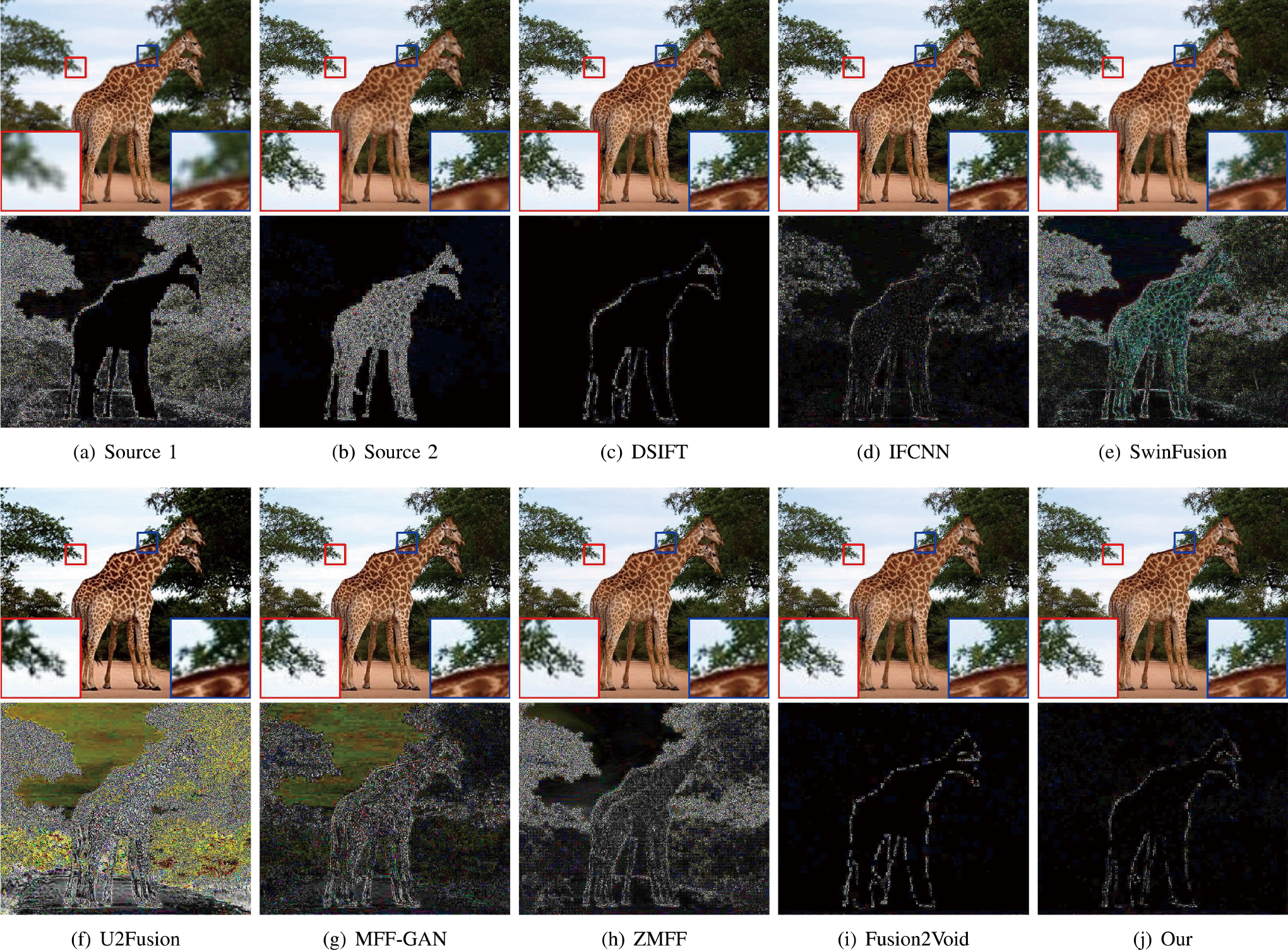}
\caption{Visual comparison of fusion results on the MFI-WHU dataset. Difference maps with respect to the ground-truth all-in-focus images are provided for reference.}
\label{fig_whu}
\end{figure*}

%		\begin{figure*}[t]
%	\centering
%	
%	\subfigure[{Source 1}]{\includegraphics[width=.19\textwidth]{tip_img/whu/hebing/25-A.png}}
%	\subfigure[{Source 2}]{\includegraphics[width=.19\textwidth]{tip_img/whu/hebing/25-B.png}}
%	\subfigure[{DSIFT}]{\includegraphics[width=.19\textwidth]{tip_img/whu/hebing/MFI-WHU_25_DSIFT.png}}
%	\subfigure[{IFCNN}]{\includegraphics[width=.19\textwidth]{tip_img/whu/hebing/MFI-WHU_25_IFCNN.png}}
%	\subfigure[{SwinFusion}]{\includegraphics[width=.19\textwidth]{tip_img/whu/hebing/swinfusion.png}}\\
%	\subfigure[{U2Fusion}]{\includegraphics[width=.19\textwidth]{tip_img/whu/hebing/U2Fusion.png}}
%	%		\vspace{-0.1in}
%	\subfigure[{MFF-GAN}]{\includegraphics[width=.19\textwidth]{tip_img/whu/hebing/MFF-GAN.png}}
%	\subfigure[{ZMFF}]{\includegraphics[width=.19\textwidth]{tip_img/whu/hebing/ZMFF.png}}
%	\subfigure[{Fusion2Void}]{\includegraphics[width=.19\textwidth]{tip_img/whu/hebing/fusion2void.png}}
%	\subfigure[{Our}]{\includegraphics[width=.19\textwidth]{tip_img/whu/hebing/ourour.png}}\\
%	
%	
%	\caption{Visual comparison of fused images on the MFFW dataset. Corresponding difference maps with respect to Source Image 1 are also presented.}
%	\label{fig_WHU}
%\end{figure*} 

\section{Experiments}
\subsection{Experimental setting}
\subsubsection{Datasets} To evaluate the effectiveness of the proposed IPS, we conduct extensive comparative experiments on four publicly available MFIF datasets. These include three real-world datasets, namely Lytro \cite{Lytro}, MFFW \cite{MFFW}, and Real-MFF \cite{zhang2020real}, as well as one synthetic dataset, MFI-WHU \cite{MFF-GAN}. Lytro is one of the most frequently used benchmarks in the MFIF domain. It contains 20 pairs of multi-focus images captured using a light field camera, with each image having a resolution of $520 \times 520$ pixels. The MFFW dataset provides 13 image pairs characterized by strong defocus spread effects, making it particularly challenging for fusion algorithms. Real-MFF is a large-scale dataset comprising 710 image pairs. Unlike Lytro and MFFW, it includes corresponding all-in-focus ground-truth images. The synthetic MFI-WHU dataset is constructed by applying a Gaussian blur to 120 natural images selected from the COCO dataset, resulting in paired multi-focus images with available ground-truth.

\subsubsection{Compared methods}
To thoroughly evaluate the performance of IPS, we compare it with nine representative state-of-the-art MFIF methods. These include traditional approaches such as DSIFT \cite{DSIFT} and DCT\_Corr\cite{DCT_Corr}, supervised deep learning techniques like CNN \cite{liu2017_supervised}, IFCNN \cite{IFCNN}, and SwinFusion \cite{Swinfusion}, as well as unsupervised learning methods like U2Fusion \cite{U2fusion}, MFF-GAN \cite{MFF-GAN}, ZMFF \cite{ZMFF}, and Fusion2Void \cite{lin2024fusion2void}.

\subsubsection{Evaluation metrics}
For the Real-MFF and MFI-WHU datasets, where ground-truth fused images are available, we adopt reference-based metrics, including Peak Signal-to-Noise Ratio (PSNR) and Structural Similarity Index Measure (SSIM), to quantify the fidelity of the fusion results.
In the cases of Lytro and MFFW, where ground-truth is unavailable, we employ several no-reference quality indicators. The normalized mutual information \(\mathcal{Q}_{MI}\) \cite{Q_MI} is used to assess how well the fused image preserves information from the source images. The spatial frequency \(\mathcal{Q}_{SF}\) \cite{Q_SF} measures the amount of high-frequency detail retained. The structural similarity metric \(\mathcal{Q}_{S}\) \cite{Q_S} evaluates structural consistency between the fused image and the source images. The perceptual quality metric \(\mathcal{Q}_{CB}\) \cite{Q_CB} reflects human visual perception. The edge-based similarity measure \(\mathcal{Q}_{AB/F}\) \cite{Q_AB/F} quantifies the preservation of edge information. The nonlinear correlation information entropy \(\mathcal{Q}_{NCIE}\) \cite{Q_NICE} captures the integrity of details, structures, color, and luminance in the fusion output.

\subsection{Comparative results}

We begin by presenting the experimental results on the Lytro and MFFW datasets. The quantitative evaluations are summarized in Tables \ref{t1} and \ref{t2}. As evident from the results, supervised deep learning methods outperform traditional techniques. Among the unsupervised approaches, only Fusion2Void delivers competitive performance, while the others yield suboptimal results. Our proposed IPS achieves the best scores on most metrics. It is also worth noting that the performance on the MFFW dataset (Table \ref{t2}) is generally lower than that on the Lytro dataset (Table \ref{t1}), reflecting the increased difficulty of fusion tasks in the MFFW dataset. Figs. \ref{fig_lytro} and \ref{fig_MFFW} provide subjective comparisons of fusion results on the Lytro and MFFW datasets, respectively. DSIFT fails to preserve fine details, especially for MFFW images where the defocus spread effect is severe. IFCNN produces unnatural transitions at the boundaries between focused and defocused regions. SwinFusion and U2Fusion introduce noticeable color distortions in the fused images. MFF-GAN and ZMFF tend to blur the results slightly. Fusion2Void sometimes generates jagged textures along edges, as illustrated in Fig. \ref{fig_lytro}. In contrast, IPS produces the highest visual quality, effectively integrating all focused pixels from the source images without introducing color or texture artifacts. Figs. \ref{fig_lytro} and \ref{fig_MFFW} also include the difference maps of each fused image relative to Source Image 1. A difference map with less residual information indicates a higher similarity between the fused image and Source Image 1. Therefore, attention should be directed to the regions in the difference maps corresponding to the focused areas of Source Image 1. IPS consistently yields the lowest residuals in these maps, suggesting its superior capability in preserving high-frequency focused details. This advantage stems from its ability to learn pixel-wise focus classification, enabling the accurate extraction of focused content from partially focused inputs. As shown in the magnified regions in Fig. \ref{fig_MFFW}, IPS better preserves subtle structures(e.g. the small red flowers), demonstrating its strong focus discrimination ability at a fine-grained level.

Further experiments are conducted on the Real-MFF and MFI-WHU datasets, where ground-truth all-in-focus images are available. Table \ref{ssimpsnr} reports the PSNR and SSIM values for all methods. As reference-based metrics, PSNR and SSIM provide more reliable and objective evaluations compared to the no-reference metrics used previously. IPS achieves the highest PSNR and SSIM on both datasets, with a notably larger PSNR margin over competing methods, indicating that its fused images are closer to the all-in-focus ground-truth. Figs. \ref{fig_realMFF} and \ref{fig_whu} present representative fused images and their corresponding difference maps with respect to the ground-truth. The source images in Fig. \ref{fig_realMFF}, taken from the Real-MFF dataset, feature challenging scenes where focused and defocused pixels are difficult to distinguish. DSIFT, SwinFusion, and U2Fusion fail to accurately locate the focused–defocused boundaries, producing overly smooth results, such as blurred building edges. MFF-GAN and ZMFF introduce color distortions, while Fusion2Void leaves residual artifacts. In contrast, IPS nearly perfectly integrates all focused pixels, yielding high-quality fused images. The source images in Fig. \ref{fig_whu} are from the synthetic MFI-WHU dataset. While most methods perform well on these images, the difference maps show that IPS is significantly more effective at preserving high-frequency details than the compared methods.

\begin{table}[]
		\centering
	\small
	\renewcommand{\tabcolsep}{15pt} 
	\renewcommand{\arraystretch}{1.7}
	\captionsetup{
		singlelinecheck=off, 
		justification=centering, 
		labelsep=newline, 
		textformat=simple,
	}
	\caption{
		{Quantitative results of the ablation study on network architecture. “w/o” denotes “without”.}}
	\begin{tabular}{c|cc}
		\hline
		Network                & PSNR  & SSIM  \\ \hline
		w/o global branch  & 36.98 & 0.965 \\ \hline
		w/o local branch & 38.83 & 0.986 \\ \hline
		Both branches retained & \textcolor{red}{42.19} & \textcolor{red}{0.991} \\ \hline
	\end{tabular}
		\label{t_ablation_net}
\end{table}

% Our IPS achieves the highest scores on both datasets, with PSNR improvements that are particularly pronounced. These results indicate that IPS produces fused images that are significantly closer to the ground-truth. 
%Figs. \ref{fig_realMFF} and \ref{fig_whu} show representative fusion results produced by different methods, along with the corresponding difference maps with respect to the ground-truth all-in-focus images. Despite variations in data distribution, IPS reliably identifies focused regions in the source images and integrates them to generate high-quality fusion outputs. The difference maps indicate that IPS retains high-frequency details more effectively than the competing approaches.

% In this section, we showcase the experimental results obtained from evaluations on three MFIF datasets: Lytro \cite{Lytro}, MFFW \cite{MFFW} and MFI-WHU \cite{MFF-GAN}, aimed at assessing the effectiveness of our proposed method. The experiments validate the proposed method's efficacy and demonstrate its superior performance over existing state-of-the-art methods across all three datasets.

\begin{figure*}[t]
	\centering
	\includegraphics[width=\textwidth]{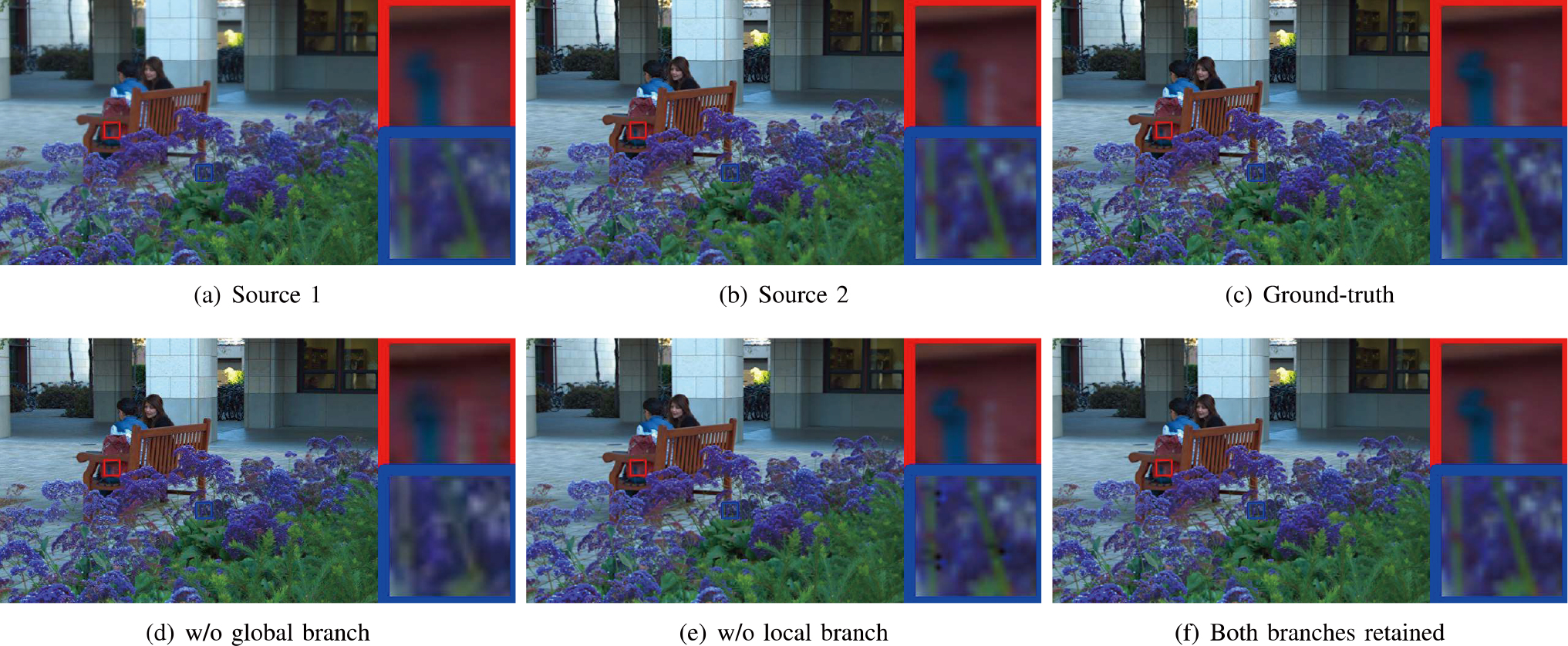}
	\caption{Visual results of network architecture ablation study.}
	\label{fig_ablation_net}
\end{figure*}

%\begin{table}[]
%	\centering
%	\small
%	\renewcommand{\tabcolsep}{15pt} 
%	\renewcommand{\arraystretch}{1.7}
%	\captionsetup{
%		singlelinecheck=off, 
%		justification=centering, 
%		labelsep=newline, 
%		textformat=simple,
%	}
%	\caption{
%		{Quantitative results of the ablation study on network architecture. “w/o” denotes “without”.}}
%	\begin{tabular}{c|cc}
%		\hline
%		Filter                & PSNR  & SSIM  \\ \hline
%		Median filter  & 41.51 & 0.987 \\ \hline
%		Gaussian filter  & 41.86 & 0.989 \\ \hline
%		Mean filter & \textcolor{red}{42.19} & \textcolor{red}{0.991} \\ \hline
%	\end{tabular}
%	\label{t_ablation_filter}
%\end{table}

\subsection{Ablation studies}
\subsubsection{Network architecture}
Our IPS adopts a cross-image fusion network as its backbone, which consists of two branches: a local feature extraction branch and a global feature extraction branch. The local branch, built with ResBlocks, captures fine-grained local features from the source images, while the global branch, implemented with Mamba blocks, models long-range dependencies among non-local focused pixels. We conduct an ablation study on the Real-MFF dataset to assess the contribution of each branch, with results presented in Table \ref{t_ablation_net} and Fig. \ref{fig_ablation_net}. As shown in Table \ref{t_ablation_net}, removing either branch degrades fusion performance in terms of PSNR and SSIM, whereas retaining both yields the best results. Without the global branch, the network fails to capture non-local focus dependencies, causing color distortions (Fig. \ref{fig_ablation_net}(d)). Without the local branch, the lack of local feature extraction leads to the loss of fine details (Fig. \ref{fig_ablation_net}(e)). With both branches, IPS effectively integrates local and non-local information, producing fused images with rich details, visual coherence, and high fidelity.
\begin{table}[]
	\centering
	\small
	\renewcommand{\tabcolsep}{15pt} 
	\renewcommand{\arraystretch}{1.7}
	\captionsetup{
		singlelinecheck=off, 
		justification=centering, 
		labelsep=newline, 
		textformat=simple,
	}
	\caption{
		{Quantitative comparison of IPS fusion performance using different filters.}}
	\begin{tabular}{c|cc}
		\hline
		Filter                & PSNR  & SSIM  \\ \hline
		Median filter  & 41.51 & 0.987 \\ \hline
		Gaussian filter  & 41.86 & 0.989 \\ \hline
		Mean filter & \textcolor{red}{42.19} & \textcolor{red}{0.991} \\ \hline
	\end{tabular}
	\label{t_ablation_filter}
\end{table}
\subsubsection{Low-pass filter}
IPS treats any optical image as an all-in-focus image and applies low-pass filtering to generate the corresponding defocused image. The filter used in IPS is a mean filter with a kernel size randomly varied within the range [3, 31]. We further investigate the impact of different filters on the fusion performance of IPS by comparing the mean filter with the median and Gaussian filters, all with kernel sizes randomly varied within the same range [3, 31]. Experiments conducted on the Real-MFF dataset are summarized in Table \ref{t_ablation_filter}. The results demonstrate that the choice of filter has only a minor impact on fusion quality. All tested filters effectively produce defocused pixels, enabling the network to learn focus–defocus pixel classification irrespective of the filter type. Notably, IPS achieves relatively better fusion performance with the mean filter, which is therefore selected for training data generation.

\begin{table}[]
	\centering
	\small
	\renewcommand{\tabcolsep}{15pt} 
	\renewcommand{\arraystretch}{1.7}
	\captionsetup{
		singlelinecheck=off, 
		justification=centering, 
		labelsep=newline, 
		textformat=simple,
	}
	\caption{
		{Quantitative comparison of IPS fusion performance using different mask ratios.}}
	\begin{tabular}{c|cc}
		\hline
	Mask ratio                & PSNR  & SSIM  \\ \hline
		$p=0.1$  & 36.19 & 0.962 \\ \hline
		$p=0.3$  & 36.26 & 0.963 \\ \hline
		$p=0.5$ & \textcolor{red}{42.19} & \textcolor{red}{0.991} \\ \hline
				$p=0.7$  & 36.28 & 0.963 \\ \hline
					$p=0.9$  & 36.14 & 0.962 \\ \hline
	\end{tabular}
	\label{t_ablation_p}
\end{table}

\subsubsection{Mask ratio $p$}

IPS employs a random mask $m$ to generate training images for pixel classification, as defined in Eq. (\ref{eq:2}). Each element in $m$ is set to zero with probability $p$, with a default value of $0.5$. To assess the influence of $p$ on fusion performance, we conducted experiments with $p \in \{0.1, 0.3, 0.5, 0.7, 0.9\}$ using the Real-MFF dataset. The results are summarized in Table \ref{t_ablation_p}. We observe that when $p=0.5$, the randomness in mask $m$ is maximized, leading to the greatest pixel-level randomness in the source images ${\tilde I}_f$ and ${\tilde I}_d$ produced by Eq. (\ref{eq:2}).
In this scenario, the network must accurately discriminate the focus status of each individual pixel to reconstruct the all-in-focus image $I_f$ from the heavily shuffled ${\tilde I}_f$ and ${\tilde I}_d$.
Consequently, IPS achieves its best fusion performance at $p=0.5$.

%The results, summarized in Table \ref{t_ablation_p}, show that $p=0.3$ and $p=0.7$ yield nearly identical fusion performance. This arises because IPS randomly swaps the positions of the source images during training, causing $I_f$ and $I_d$  in Eq. (\ref{eq:2}) to interchange. As a result, $p=0.3$ and $p=0.7$ produce equivalent training data and thus similar performance. The same complementarity holds for $p=0.1$ and $p=0.9$. Notably, $p=0.5$ results in the maximum number of pixel swaps between $I_f$ and $I_d$, requiring the network to distinguish focus at the pixel level to reconstruct an all-in-focus image. Consequently, $p=0.5$ delivers the best overall fusion performance.

\section{Conclusion}
This paper introduces IPS, a novel multi-focus image fusion framework that reformulates the task as a focused–defocused pixel classification problem. Training data are synthesized by randomly swapping pixels at identical spatial locations between arbitrary optical images and their blurred counterparts, allowing the network to learn multi-focus fusion without relying on any multi-focus images during training. This strategy effectively mitigates the dependency of deep learning-based MFIF methods on large-scale labeled datasets, thereby enhancing their practicality in data-scarce domains such as remote sensing and microscopy. Extensive experiments demonstrate that IPS consistently outperforms existing deep learning approaches in fusion quality, highlighting its potential as a robust and widely applicable solution.

\bibliographystyle{IEEEtran}
\bibliography{IEEEabrv,lite}

\end{document}